\documentclass[conference]{IEEEtran}
\IEEEoverridecommandlockouts
\usepackage{cite}
\usepackage{amsmath,amssymb,amsfonts}
\usepackage{algorithmic}
\usepackage{graphicx}
\usepackage{textcomp}
\usepackage{xcolor}
\def\BibTeX{{\rm B\kern-.05em{\sc i\kern-.025em b}\kern-.08em
    T\kern-.1667em\lower.7ex\hbox{E}\kern-.125emX}}

\usepackage{caption}
\usepackage{subcaption}

\usepackage{acronym}
\acrodef{dos}[DOS]{Denial-of-Service}
\acrodef{dop}[DOP]{Denial-of-Prediction}
\acrodef{eot}[EOT]{Expectation Over Transformation}
\acrodef{fgsm}[FGSM]{Fast Gradient Sign Method}
\acrodef{pgd}[PGD]{Projected Gradient Descent}
\acrodef{art}[ART]{Adversarial Robustness Toolbox}
\acrodef{red}[RED]{Reverse Engineering Deception}
\acrodef{gan}[GAN]{Generative Adversarial Network}
\acrodef{aml}[AdvML]{Adversarial Machine Learning}
\acrodef{lstm}[LSTM]{long short-term memory}
\acrodef{ai}[AI]{Artificial Intelligence}
\acrodef{ml}[ML]{Machine Learning}
\acrodef{gard}[GARD]{Guaranteeing AI Robustness against Deception}
\acrodef{ili}[ILI]{Influenza-Like Illness}
\acrodef{fpr}[FPR]{false positive rate}
\acrodef{tpr}[TPR]{true positive rate}
\acrodef{auc}[AUC]{area under the ROC curve}
\acrodef{cnn}[CNN]{convolutional neural network}
\acrodef{soa}[SOA]{state-of-the-art}
\acrodef{cs}[CS]{compressed sensing}
\acrodef{tdd}[TDD]{Task Description Document}
\acrodef{nwpu}[NWPU-RESISC45]{Northwestern Polytechnical University Remote
Sensing Image Scene Classification dataset}
\acrodef{gis}[GIS]{geographic information system}
\acrodef{ucf}[UCF101]{UCF 101 action dataset}

\usepackage{url}

\begin{document}

\title{Reverse engineering adversarial attacks with fingerprints from adversarial examples\\
\thanks{This material is based upon work supported by the Defense Advanced Research Projects Agency (DARPA) under Agreement No. HR00112090134. Approved for public release; distribution is unlimited.}
}

\author{\IEEEauthorblockN{1\textsuperscript{st} David Nicholson}
\IEEEauthorblockA{\textit{Embedded Intelligence} \\
Gaithersburg, MD, United States of America \\
david@embedintel.com}
\and
\IEEEauthorblockN{2\textsuperscript{nd} Vincent Emanuele}
\IEEEauthorblockA{\textit{Embedded Intelligence} \\
Gaithersburg, MD, United States of America \\
vince@embedintel.com}
}

\maketitle

\begin{abstract}
In spite of intense research efforts, 
deep neural networks remain vulnerable to adversarial examples: 
an input that forces the network to confidently produce incorrect outputs. 
Adversarial examples are typically generated by an attack algorithm
that optimizes a perturbation added to a benign input. 
Many such algorithms have been developed. 
If it were possible to reverse engineer attack algorithms from adversarial examples, 
this could deter bad actors because of the possibility of attribution. 
Here we formulate reverse engineering as a supervised learning problem 
where the goal is to assign an adversarial example 
to a class that represents the algorithm and parameters used. 
To our knowledge it has not been previously shown whether this is even possible. 
We first test whether we can classify the perturbations added to images by attacks 
on undefended single-label image classification models. 
Taking a ``fight fire with fire'' approach, 
we leverage the sensitivity of deep neural networks to adversarial examples, 
training them to classify these perturbations. 
On a 17-class dataset (5 attacks, 4 bounded with 4 epsilon values each), 
we achieve an accuracy of 99.4\% with a ResNet50 model trained on the perturbations.
We then ask whether we can perform this task 
without access to the perturbations, 
obtaining an estimate of them with signal processing algorithms, 
an approach we call ``fingerprinting''. 
We find the JPEG algorithm serves as a simple yet effective fingerprinter (85.05\% accuracy), 
providing a strong baseline for future work. 
We discuss how our approach can be extended to attack agnostic, learnable fingerprints, 
and to open-world scenarios with unknown attacks.

\end{abstract}


\section{Introduction}
\label{sec:intro}

Deep neural networks are susceptible to adversarial examples \cite{szegedyIntriguingPropertiesNeural2014,goodfellowExplainingHarnessingAdversarial2015}: 
inputs optimized to produce incorrect or unexpected outputs. 
Typically adversarial samples are generated by 
optimizing a perturbation $\mathbf{\delta}$ added to a benign image $\mathbf{x}$ \cite{madryDeepLearningModels2019}. 
This added perturbation can be optimized by one of an ever-growing list of attack algorithms 
\cite{tabassiTaxonomyTerminologyAdversarial2019,akhtarAdvancesAdversarialAttacks2021},  
e.g., by maximizing the loss of the softmax function 
used to train single-label neural networks for image classification.

It remains unclear whether it will prove 
too computationally expensive 
or theoretically impossible 
\cite{goodfellowExplainingHarnessingAdversarial2015,warde201611,tanayBoundaryTiltingPersepective2016,shafahiAreAdversarialExamples2020,yangBoundaryThicknessRobustness2021,daiImageClassifiersCan2021,yousefzadehDecisionBoundariesConvex2022}
to completely defend neural networks from adversarial attacks, 
at least for neural network models in their current mathematical formulation.
Defenses are notoriously difficult to evaluate, 
in spite of concerted efforts by the community to establish good practices \cite{carliniEvaluatingAdversarialRobustness2019}. 
Given the difficulties faced in developing defenses against adversarial attacks, 
we consider a different question. 
We ask whether it is possible to reverse engineer adversarial attacks, 
using adversarial examples. 
If this were possible, 
it could deter bad actors from deploying adversarial example in the real world, 
e.g., due to the threat of attribution. 
Accordingly, we identified two capabilities that would be 
desirable in a system for classifying and searching datasets of adversarial examples. 
We describe these two capabilities then 
explain how we formulate a machine learning task that can produce 
models with those capabilities.

\begin{figure}[t]
\includegraphics[width=0.95\columnwidth]{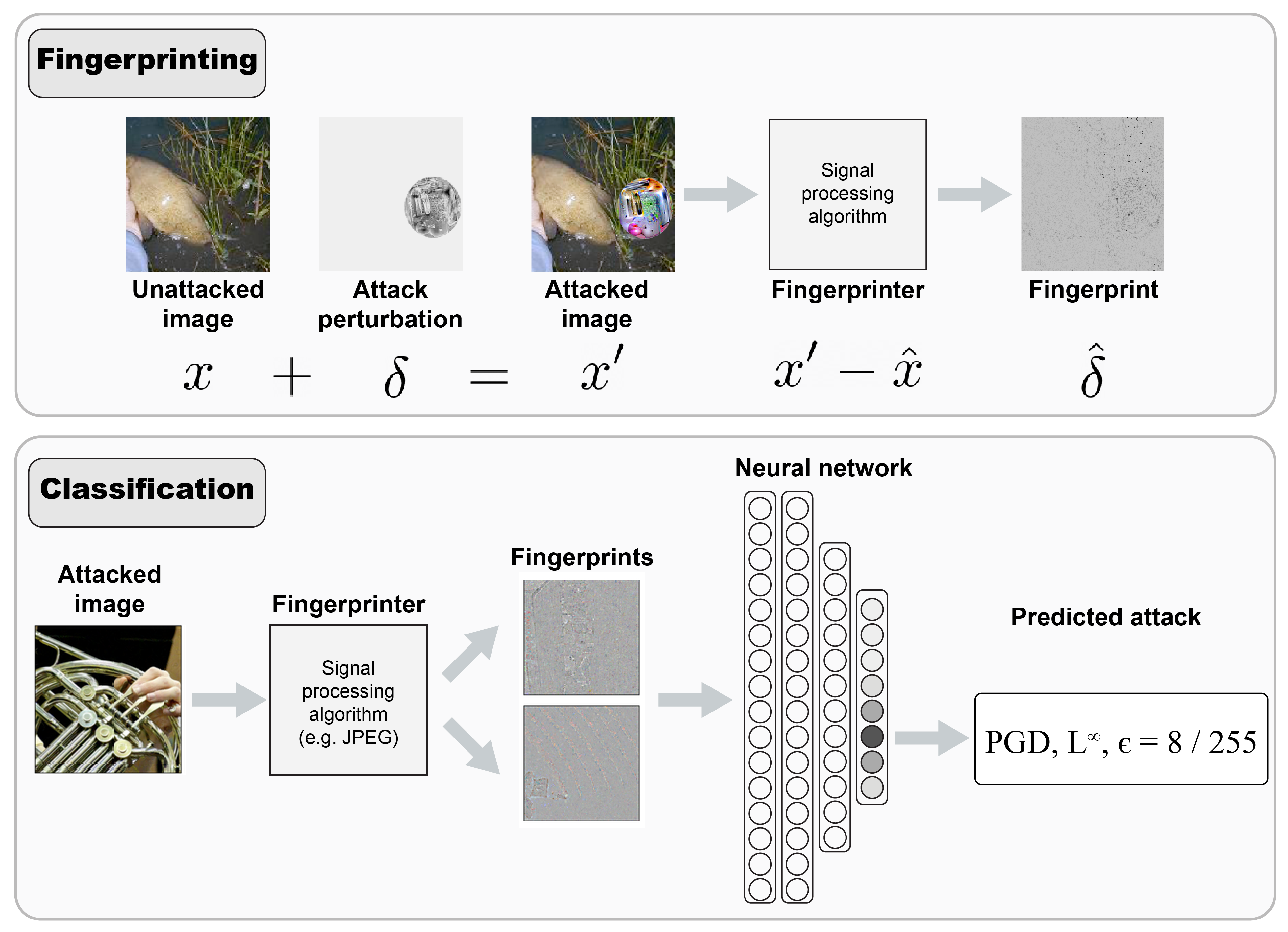}
\caption{Schematic of our approach to reverse engineering adversarial attacks.
Top row illustrates ``fingerprinting'' adversarial examples 
to obtain an estimate of the perturbation $\delta$ added by an attack 
to an image.
Bottom row illustrates training deep neural networks to classify fingerprints 
from adversarial examples, 
assigning each to a class that corresponds to an 
attack algorithm and parameters. 
Our goal in formulating the problem this way is to 
train models on large datasets of fingerprints from attacks generated 
with existing software frameworks, 
and then classify attacked images without knowledge of the attack 
or access to the perturbation, by using fingerprints.
}
\label{fig:approach-schematic}
\end{figure}

\subsection{Classifying perturbations by attack}

The first capability we would want such a system to have is 
to classify adversarial examples by 
the attack algorithm used to generate them. 
Of course it may be possible to reverse engineer attacks 
without classifying them, 
e.g., by literally "reversing" the optimization. 
In other words, here we formulate reverse engineering as 
a supervised classification problem.
Our motivation is pragmatic. 
We ask whether we can leverage widely used and well understood methods, 
as well as software frameworks that have been developed 
to generate large datasets of adversarial examples, 
that we can use to train a machine learning model 
to classify by attack algorithm. 

To the best of our knowledge, it is still an open question in the literature 
whether adversarial examples can be classified by attack algorithm. 
We emphasize that classifying examples by attack algorithm is different 
from detecting that an image has been attacked, 
e.g. by monitoring the outputs of the model for outliers, 
or (equivalently) classifying an image as an adversarial example \cite{carliniAdversarialExamplesAre2017a}. 

It could be the case that very different attack algorithms all arrive at similar perturbations, 
in which case it would be difficult or impossible to classify them.
For example, a PGD attack with an $L^\infty$ bound of $\epsilon$=8/255 
on a specific instance of a ResNet50 model
may produce perturbations that are indistinguishable 
from those produced by a Square attack with the same bound, on the same model.
Alternatively, it could be the case that, 
because of the abundance of adversarial examples in image space, 
each algorithm can produce unique perturbations that other attacks are much less likely to find. 
By the same token, 
it could be the case that classifying adversarial examples by attack algorithm 
can be done using the attacked images themselves, 
and therefore is trivial. 
Alternatively, classification of attacks might require the perturbation added by an attacker, 
which a defender may not have access to, even when they know the image is attacked. 
These open questions make it important to carry out a rigorous study 
of whether or not this task is even possible.

\subsubsection{Classifying families of adversarial attacks}

A related question that arises when considering how to reverse engineer adversarial attacks 
is whether this problem is hierarchical. 
Can attacks be grouped somehow, and could this grouping help with classification?
In this work we again operationally define families of attacks 
with the goal of understanding how this relates to our ability to classify them. 
For brevity we avoid summarizing the history of adversarial attacks, 
and refer the reader to recent reviews.
These reviews show that attacks  can be schematized in several ways 
\cite{tabassiTaxonomyTerminologyAdversarial2019,akhtarAdvancesAdversarialAttacks2021}, 
and grouped along several dimensions. 
Although it is important to consider all dimensions, 
in our studies we focus on two: the \textit{threat model}, and the \textit{constraint}. 
Researchers in this area often speak of a \textit{threat model}, 
a term that summarizes the access that an attacker has to the machine learning model. 
Under a \textit{white-box} threat model, 
the attacker has total access, 
and thus can compute the gradient, 
which generally speaking allows for more powerful attacks. 
In contrast, under a \textit{black-box} threat model, 
the attacker only has access to outputs of the machine learning model, 
which may be a single predicted label or a vector of scores. 
Black-box attacks generally require many more queries 
or iterations to produce a powerful attack. 
Another dimension along which attacks can be grouped 
is the constraints placed on the perturbation. 
It is common for both white-box and black-box attacks to use $L^p$-ball constraints, 
where the size of a perturbation generated by an attack 
is constrained to be less than some $\epsilon$ as computed with the corresponding $p-\text{norm}$ 
(e.g., an $L^2$ norm).
In contrast, patch attacks do not constrain the perturbation size 
but constrain the attack to a confined space within the image \cite{brown2017adversarial}.
Considering just these two dimensions, 
we can already begin to group each attack algorithm into a family. 
E.g., 
both \ac{fgsm}\cite{goodfellowExplainingHarnessingAdversarial2015} 
and \ac{pgd}\cite{madryDeepLearningModels2019},
could be considered 
part of a family of white-box, $L^p$-ball attacks, 
while square attack \cite{andriushchenkoSquareAttackQueryefficient2020} 
could be considered part of the family of black-box, $L^p$-ball attacks. 
We of course recognize that attacks can be grouped in other ways
---e.g., one might consider a patch attack a physical attack on real-world objects, 
as opposed to white-box attacks on images\cite{akhtarThreatAdversarialAttacks2018}---
but we think most researchers would not find it controversial to group attacks into families, 
and agree that it would be useful to ask 
whether the ability to classify adversarial examples depends in part on the family. 
An effect of family figures into our analyses below.

\subsection{Estimating perturbations in an attack-agnostic manner}

Assume for a moment that we can classify adversarial examples by attack algorithm, 
but that this is best done using the perturbation added by the attack, 
not the attacked image itself (as our result below indicate). 
This suggests that the second capability we will want our system to have 
is to estimate the true perturbation $\mathbf{\delta}$ generated by an attack, 
ideally in an attack-agnostic fashion 
that does not require us to develop a reverse engineering method 
for each algorithm or family of attacks. 
Here again, little work has been done, 
and so we take a methodical approach.
As we detail in Section~\ref{sec:approach}, 
we test whether we can obtain estimates of the perturbations 
using familiar signal-processing methods 
for image compression and restoration, 
namely the JPEG algorithm and a compressed sensing-based image reconstruction algorithm. 
We were first motivated to take this approach 
after observing that simply reconstructing attacked images $\mathbf{x'}$
and subtracting the reconstruction $\mathbf{\hat{x}}$ from $\mathbf{x'}$
to obtain an estimate of the perturbation $\mathbf{\hat{\delta}}$
can produce a mark that is obvious upon visual inspection. 
We show an example in the top panel of Figure~\ref{fig:approach-schematic}.
We refer to the estimates of $\mathbf{\hat{\delta}}$ so obtained as "fingerprints". 
Our motivation for this approach also springs from 
previous work showing that such algorithms 
can remove the perturbation $\mathbf{\delta}$ added by an attacker \cite{guoCounteringAdversarialImages2018}.
We are of course aware of work showing that adaptive attacks 
can produce perturbations that successfully attack models even after input transformations 
such as JPEG are applied \cite{shin2017jpeg}.
Our express goal here is \textit{not} to defend the model, 
but to obtain an estimate of the perturbation 
in an attack-agnostic fashion. 
We see the simplifying assumption of 
testing on undefended models 
as part of our methodical approach, 
and consider it important to test in this simplified setting 
to lay the foundation for future work. 

\subsection{Our contribution}

Without a theory of adversarial examples, 
we cannot state unequivocally that each capability can be designed, 
but we can test whether each is possible empirically. 
Here we provide evidence that such a system can be designed 
with the two capabilities just described, 
suggesting it will be possible to reverse engineer adversarial attacks. 
Our contributions are as follows:
\begin{itemize}
    \item We show that given the true perturbations $\delta$ added to benign images, 
        we can predict with near perfect accuracy the attack used 
        and at least one of its parameters, the value of the bound $\epsilon$ for bounded attacks.
    \item We then show that we can obtain estimates of the perturbation, 
        which we call fingerprints, in an attack agnostic manner 
        that still allows us to classify the perturbations according to attack.
    \item We demonstrate that fingerprints obtained with simple signal processing methods 
        allow us to classify attacks with accuracy of 84.49\%.
        Compared to our empirical upper bound of near 100\% accuracy 
        shown with true perturbations, this leaves room for improvement, 
        but we provide a strong baseline for future work using data-driven methods, 
        a point we return to in the discussion.
\end{itemize}

\section{Approach}
\label{sec:approach}

\subsection{Notation}

To discuss our method, we adopt the following notation:  
Let $F_{\theta}$ be a deep neural network model
with parameters $\theta$ trained to map 
inputs $\mathbf{x}$ to 
to a set of $c$ class labels $Y = \{y_1, y_2, ..., y_c\}$, 
For all supervised learning problems, 
we update $\theta$ by minimizing a loss function $L$.

Where needed, we use a superscript $F^{\hat{\delta}}_{\theta}$ 
to distinguish a deep neural network 
trained  to classify adversarial examples 
into attack classes $Y^{\hat{\delta}} = \{y_1, y_2, ..., y_c\}$ 
from the standard network for single-label image classification $F_{\theta}$. 
For such a network $F^{\hat{\delta}}_{\theta}$, 
each class $c$ in $Y$ is a specific attack algorithm from one of the families we study, 
where the class also denotes a bound and an epsilon parameter when the attack uses such constraints: 
e.g., the \ac{pgd} attack from the white-box $L^p$-ball family, 
with an $L^\infty$ bound and $\epsilon = 4 / 255$.
(We define these further below.)
Using $\hat{\delta}$ as a superscript 
denotes that we have trained $F^{\hat{\delta}}_{\theta}$ 
on some estimate of the perturbation $\mathbf{\delta}$ added by an attack to 
an image $\mathbf{x}$ to produce the adversarial example $\mathbf{x'}$. 
We call these estimates $\hat{\delta}$ ``fingerprints''. 

\subsection{Classifying adversarial images by attack algorithm}

Here we focus on attacks on deep neural network models for single-label image classification, 
as this is where much of the research on attacks has centered. 
To simplify the problem, we assume that attacked models are undefended, 
and that a researcher is able to train machine learning models on datasets of attacked images 
without threat of adversarial attack. 
Taking a ``fight fire with fire'' approach, 
we train neural networks to classify adversarial examples by the attack used 
to optimize the perturbation $\mathbf{\delta}$ added to the benign image $\mathbf{x}$.
This approach is motivated by previous work 
showing that deep neural networks are susceptible to adversarial examples 
in part because they latch on to high-frequency components of images 
that are largely imperceptible to humans 
\cite{wangHighFrequencyComponentHelps2020,yinFourierPerspectiveModel2020,zhangUniversalAdversarialPerturbations2021}.

\subsection{Adversarial attacks}

\subsubsection{Attack families, algorithms, and parameters}

Although research on adversarial attacks is rapidly evolving, 
the community of researchers has recognized families of attacks. 
As stated in Section~\ref{sec:approach}, 
we group attack algorithms into three families for our problem formulation: 
white box $L^p$-ball attacks, black box $L^p$-ball attacks, and patch attacks. 
We see this is a reasonable choice given that much work has focused on attacks 
from these three families.

As shown in Table~\ref{tab:attacks}, we generated attacks with 
two algorithms from the white box family, 
\ac{fgsm}\cite{goodfellowExplainingHarnessingAdversarial2015} 
and \ac{pgd}\cite{madryDeepLearningModels2019},
and one each from the black box and patch families: 
square attack \cite{andriushchenkoSquareAttackQueryefficient2020}
and the universal patch attack of \cite{brown2017adversarial}, 
respectively.
For purposes of classification, 
we further divided attacks by the value of $p$ 
used for the $L^p$ norm, 
(i.e., $L^2$ or $L^\infty$)
and the value of the bound parameter $\epsilon$, 
where we used 4 unique values per attack and norm. 
In total this gave us 17 different classes, 
as shown in Table~\ref{tab:attacks}.

\begin{table}
\begin{tabular}{llll}
\textbf{Attack algorithm} & \textbf{Attack family}                                         & \textbf{$\epsilon$ values}                                         & \textbf{Other parameters}                                                                        \\
FGSM                      & \begin{tabular}[c]{@{}l@{}}White box, \\ $L^p$-ball\end{tabular} & \begin{tabular}[c]{@{}l@{}}\{1, 2, 4, 8\} \\ / 255\end{tabular}    & None                                                                                             \\
PGD, $L^\infty$           & \begin{tabular}[c]{@{}l@{}}White box, \\ $L^p$-ball\end{tabular} & \begin{tabular}[c]{@{}l@{}}\{1, 2, 4, 8\} \\ / 255\end{tabular}    & \begin{tabular}[c]{@{}l@{}}250 steps,\\ step size = \\ (2.5 * $\epsilon$)\\ / steps\end{tabular}  \\
PGD, $L^2$                   & \begin{tabular}[c]{@{}l@{}}White-box, \\ $L^p$-ball\end{tabular} & \begin{tabular}[c]{@{}l@{}}\{0.25, 0.5, \\ 1.0, 2.0\}\end{tabular} & \begin{tabular}[c]{@{}l@{}}250 steps, \\ step size = \\ (2.5 * $\epsilon$)\\ / steps\end{tabular} \\
Square, $L^\infty$        & \begin{tabular}[c]{@{}l@{}}Black box, \\ $L^p$-ball\end{tabular} & \begin{tabular}[c]{@{}l@{}}\{1, 2, 4, 8\} \\ / 255\end{tabular}    & 10k queries                                                                                        \\
Universal patch           & Patch attack                                                   & None                                                               & None                                                                                            
\end{tabular}
\caption{Attack algorithms, families, and parameters used.}
\label{tab:attacks}
\end{table}

\subsubsection{Dataset}

All attacks were generated on images from 
Imagenette\footnote{\url{https://github.com/fastai/imagenette}},
a version of the ImageNet dataset with only 10 classes,
and approximately 2000 images per class.
We preserved the training and test sets from Imagenette
to avoid contaminating our test set with training images,
as explained further below.
Thus, we generated attacked images for both the training
and test sets. 
For all experiments here, we used only successful attacks. 
The number of attacked images thus generated for each
combination of attack and epsilon size (for bounded attacks) ranged
from 5750-6700 for the training set, and from 2400-2990 for the validation set.

\subsection{Taking ``fingerprints'' of perturbations}
\label{subsec:approach-prints}

After generating these pools of attacked images, we ``took fingerprints''
from them.
The same pipeline was used for all combinations of fingerprint methods and parameter settings reported.

To extract fingerprints form attacked images, we used different methods for 
image compression and reconstruction, under the working hypothesis that these methods  
will tend to remove the adversarial perturbation added to an image.
If this hypothesis is true, then we should be able to subtract the reconstructed 
image $\mathbf{\hat{x}}$ from the attacked image $x$ to obtain an estimate of 
the perturbation $\mathbf{\hat{\delta}}$ added by an attacker.  
Using this notation, for all reconstruction methods, we obtain our fingerprint like so: 

$$\mathbf{\hat\delta} = \mathbf{x'} - \mathbf{\hat{x}}$$

\paragraph{JPEG}
To use JPEG as a fingerprint extraction technique, 
we simply set the quality parameter of JPEG
and then used the compressed image $x_{jpeg}$ to create a fingerprint
$\delta_{jpeg} = x' - x_{jpeg}$.

\paragraph{Compressed sensing}
At a high level, compressed sensing is a family of algorithms 
that obtains high-fidelity estimates of a signal given many fewer samples than 
required by classical signal processing theorems, 
by solving an undetermined system of linear equations with a sparsity constraint.
The system of equations typically consists of a random sampling matrix $S$, 
a dictionary $D$ that transforms the signal into a domain where it can be considered 
sparse (e.g., DCT), and a regularization constant $\lambda$ that enforces sparsity. 
The algorithm we use also adds a $k/n$ parameter, the ratio of random samples 
to the number of true samples in the signal (for images, the percentage of pixels). 
By randomly grabbing a subset of samples, in effect we use a Bernoulli sampling matrix.

Given this set of parameters for compressed sensing, $(S,D,k/n,\lambda)$, we solve for the attack
fingerprint $\delta_{cs}$ as
\begin{align}
  \chi_{cs}=\left\{\chi \in {\rm I\!R}^n | \min_\chi\|b-SD\chi\|_2 + \lambda\|\chi\|_1\right\} \label{eq:lasso}\\
    x_{cs} = D\chi_{cs} \\
    \delta_{cs} = x' - x_{cs}.
\end{align}
\ref{eq:lasso} is typically solved using techniques described in \cite{taoLocalLinearConvergence2016}. 

\subsection{Neural network training}

\subsubsection{Dataset preparation}

We built training and test sets from our database of fingerprints 
to train and test neural network models $F^{\hat{\delta}}_\theta$ 
that assign labels to adversarial examples according to attack class. 
So that we could avoid contaminating the test set with the training set,
we maintained the original splits from Imagenette.
That is, we built our training sets using fingerprints extracted
from attacked images generated with the Imagenette training set, 
and likewise built our test sets with fingerprints 
extracted from attacked images generated with the Imagenette test set.
Both our training and test sets contained fingerprints taken from 1000 unique
images from the original Imagenette dataset.
These unique images were sampled randomly when creating the splits
from the larger pools of fingerprints generated as described above.
The training set was further split into training and validation sets,
with 90 percent of the samples used for training,
and the other 10 percent used
to validate performance during training.
Before creating splits as just described, 
we filtered the total set of adversarial examples 
to keep only successful attacks. 
For the 17-class dataset used for our main result, 
this gave us a training set size of 124156 samples 
and a test set size of 52283 samples, 
for each fingerprint or other input used to train networks 
(the true perturbation $\delta$ or the adversarial example itself).
(I.e., there was a training set of 124156 adversarial examples, 
and a separate training set of 124156 JPEG reconstructions, etc.)

\subsubsection{Model, optimizer, hyperparameters}

For all experiments using fingerprints or comparison training data, 
we used the ResNet50 architecture \cite{heDeepResidualLearning2015} 
as the neural network model $F^{\hat{\delta}}_\theta$. 
As our loss function $L$ we used standard cross-entropy loss,  
and we optimized parameters with the Adam optimizer \cite{kingma2014adam}
with the learning rate $\alpha=0.01$, and a batch size of 128. 
We configured training such that networks would train for a maximum of 50 epochs
(where each epoch is an iteration through the entire training set), but used an
early stopping scheme. Early stopping depended on accuracy as measured
on the validation set every 400 steps (i.e., every 400 batches).
If four validation checkpoints elapsed without accuracy increasing beyond
the maximum recorded, then training stopped.
This meant that in practice the optimization rarely ran
for the full 50 epochs.
Visually inspecting the training histories showed that this scheme  
enabled sufficient training for the optimization to converge
while preventing networks from overfitting on the training set.
For all experiments, we trained four replicates of ResNet50,  
where each replicate had weights initialized before training.

\section{Results}

\subsection{Classification of adversarial examples by attack algorithm}

\begin{figure*}
     \centering
     \begin{subfigure}[b]{0.375\textwidth}
         \centering
         \includegraphics[scale=0.5]{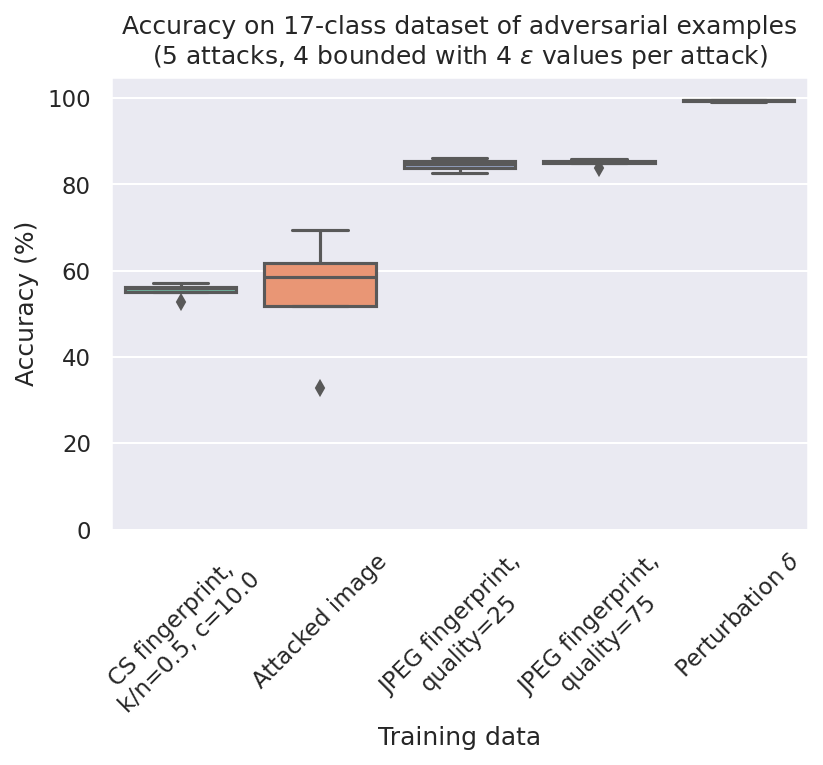}
         \caption{Accuracy of ResNet50 models trained to classify adversarial examples according to attack class.
         Box and whisker plots show distribution of accuracies across four training replicates.
         X-axis indicates type of data that models were trained to classify: either the perturbation $\delta$ 
         added by an attack to an image, the attacked image itself, or a "fingerprint", an estimate of $\delta$ 
         obtained with a signal processing algorithm.}
         \label{fig:acc-by-fingerprint}
     \end{subfigure}
     \hfill
     \begin{subfigure}[b]{0.575\textwidth}
         \centering
         \includegraphics[scale=0.15]{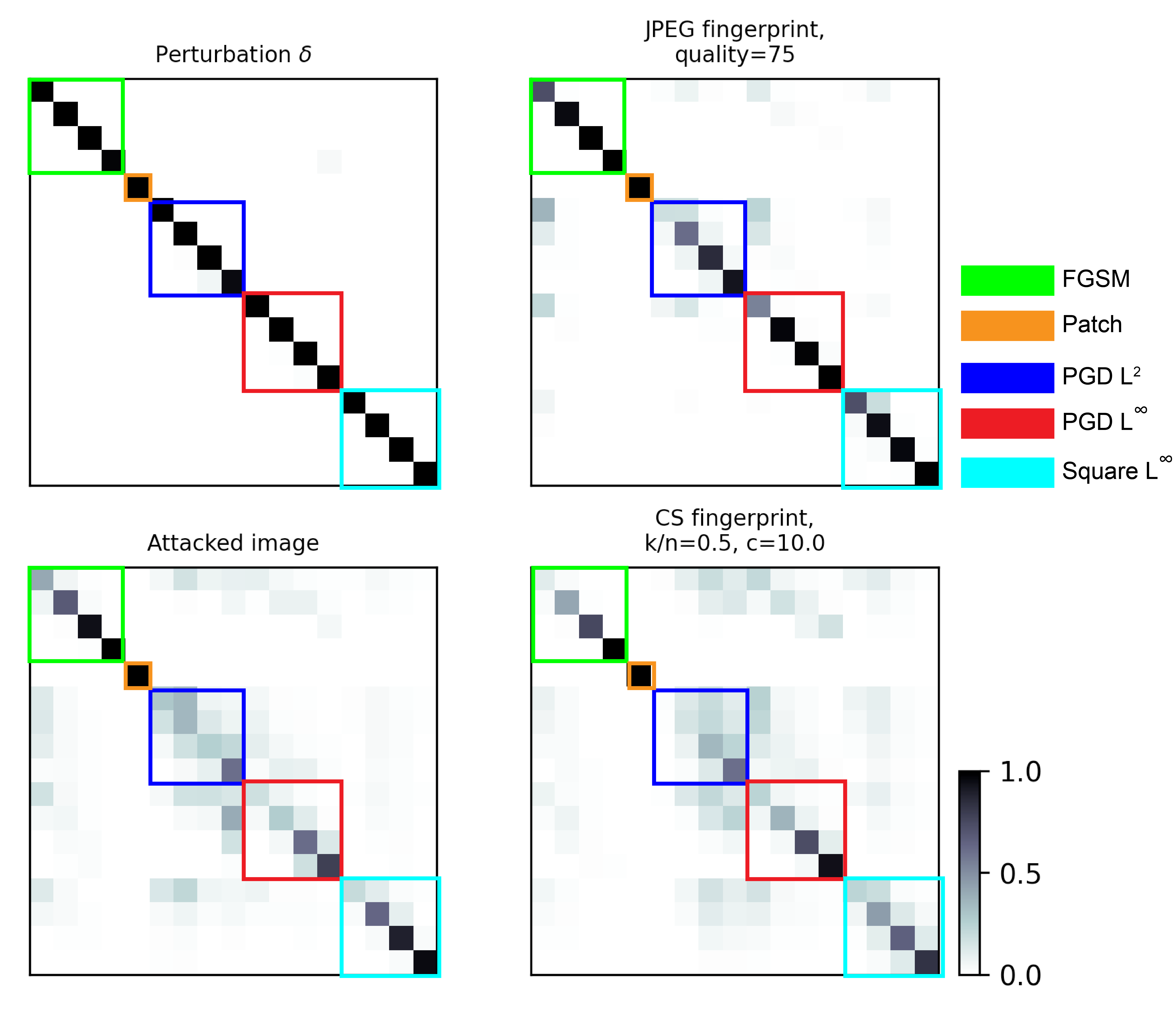}
         \caption{Confusion matrices for ResNet50 models, 
         where rows are ground truth labels, columns are predicted labels, 
         and grayscale intensity in each square indicates probability of predicting each label, normalized within rows. Colored rectangles indicate attack algorithms.
         $\epsilon$ values are sorted within attack algorithm to increase from 
         top to bottom and from left to right.
         Each plot is generated from predictions of one training replicate of a ResNet50 model. 
         In general, results were similar across replicates and so a representative example is shown.}
         \label{fig:confmat-1}
     \end{subfigure}
        \caption{Results of training deep neural networks to assign attack classes to adversarial examples}
        \label{fig:classification}
\end{figure*}

We began by asking whether it is even possible to 
classify adversarial examples by  the attack algorithm and parameters used. 
As stated in Section~\ref{sec:intro}, 
we start here because we formulate reverse engineering attacks as a supervised learning problem,  
and because to our knowledge this remains an unaddressed question in the literature. 
To answer this question, we began by training a ResNet50 on the perturbations added to images 
from the 10-class Imagenette dataset, 
by attacking a separate, undefended ResNet50 model pre-trained on all of ImageNet. 
We generated adversarial examples with 5 different attacks, 4 of them bounded, with 4 epsilon values per attack (see Section \ref{sec:approach}), giving us us a 17-class dataset. Results are shown in Figure~\ref{fig:classification}

We found that, yes, we were able to assign labels to perturbations corresponding 
to the attack algorithm and size of the epsilon bound on attacks, 
as shown in the rightmost column of Figure~\ref{fig:acc-by-fingerprint}. 
The ResNet50 trained on the perturbations alone was able to perform this task with near-perfect accuracy: 
99.4\% $\pm$ 0.15\% (mean $\pm$ standard deviation) across 4 training replicates 
(instances of a model trained from randomly initialized weights). 
In contrast, when we asked the same ResNet50 model 
to perform the same task given the attacked images themselves 
(i.e., the benign image $\mathbf{x}$ + the perturbation $\mathbf{\delta}$ we classified before), 
we were only able to achieve 54.84\% $\pm$ 15.62 \% accuracy on the held-out test set 
(Figure~\ref{fig:acc-by-fingerprint}, second column from left). 
We analyze these two results further below, but note that 
taken together they indicate that it is possible to classify perturbations 
by attack algorithm and parameters, given the true perturbation, 
and additionally suggest that it will not be sufficient to simply classify the attacked images themselves. 

\subsection{Obtaining and classifying estimates of perturbations by fingerprinting with signal-processing methods}

Given this initial evidence that it is possible to classify perturbations by attack and parameters used, 
we next asked whether we would be able to classify attacks 
even if we did not have access to the true perturbations. 
This would be the case if we detected that the image was attacked, 
by inspecting the image and comparing the human label 
with the incorrect outputs of an image-classification model, 
but we did not have knowledge of the attack used. 
In this situation, we would somehow need to obtain an estimate of the perturbation 
added by an attacker.

Motivated by previous work on defenses showing 
signal processing algorithms for image compression and reconstruction 
can remove perturbations added by 
non-adaptive adversarial attacks, 
we chose to test two of these algorithms 
as methods for obtaining estimates of the perturbation $\delta$ 
added to attacked images,
as detailed in Section~\ref{sec:approach}.

We began by testing with the JPEG algorithm.
We trained ResNet50 models on "fingerprints" produced 
by first compressing and then decompressing the attacked images with JPEG, 
treating this as an estimate of the benign image before attack $\mathbf{\hat{x}}$ 
that we subtracted from the attacked image $\mathbf{x}$ 
to give us an estimate of the perturbation added by the attack, $\mathbf{\hat{\delta}}$. 
To test for any effect of JPEG parameters, 
we generated these ``fingerprints'' with a quality parameter of 75,
as used in the original paper proposing JPEG as a defense \cite{guoCounteringAdversarialImages2018}, 
and also with quality=25. 
ResNet50 models trained on JPEG quality=75 achieved an accuracy of 
85.05\% $\pm$ 0.83 \% 
and those trained on JPEG quality=25 
achieved an accuracy of 84.49\% $\pm$ 1.46 \%. 
By comparison, models trained on compressed sensing (CS) fingerprints 
achieved 55.39\% $\pm$ 1.87\% accuracy.
These results are also shown in the middle columns of Figure~\ref{fig:acc-by-fingerprint}.

To better understand these results, 
 we generated confusion matrices for each of the models, 
and visually inspected these to see if they provided additional insights. 
These are shown in Figure~\ref{fig:confmat-1}. 
The first thing we noticed when inspecting these plots 
was the inverse relationship between the size of the epsilon bound 
and the size of the error. 
I.e., attacks generated with smaller epsilon bounds 
were more difficult to classify. 
Additionally we observed that 
attacks with an $L^\infty$ bound 
appeared to be easier to classify; 
results were consistent for these attacks 
except for the smallest epsilon values, 
whereas the networks tended to make more mistakes 
for the PGD $L^2$ attacks.

\subsection{Performance when adding more, smaller $\epsilon$ values}

Because we noted that our purely supervised classification approach 
was challenged by smaller values of $\epsilon$, 
as can be seen in the confusion matrices in \ref{fig:confmat-1}, 
and because PGD $L^2$ attacks in particular appeared to be challenging to classify, 
we chose to push on this result further. 
We generated additional PGD-$L^2$ attacks 
with another set of values: 0.1, 0.2, 0.3 and 0.4, 
and then used this expanded dataset to repeat the experiments where 
we trained ResNet50 models to classify the true perturbation $\mathbf{\delta}$ 
and the JPEG-based fingerprints. 

\begin{figure}
\centering
\includegraphics[width=0.95\columnwidth]{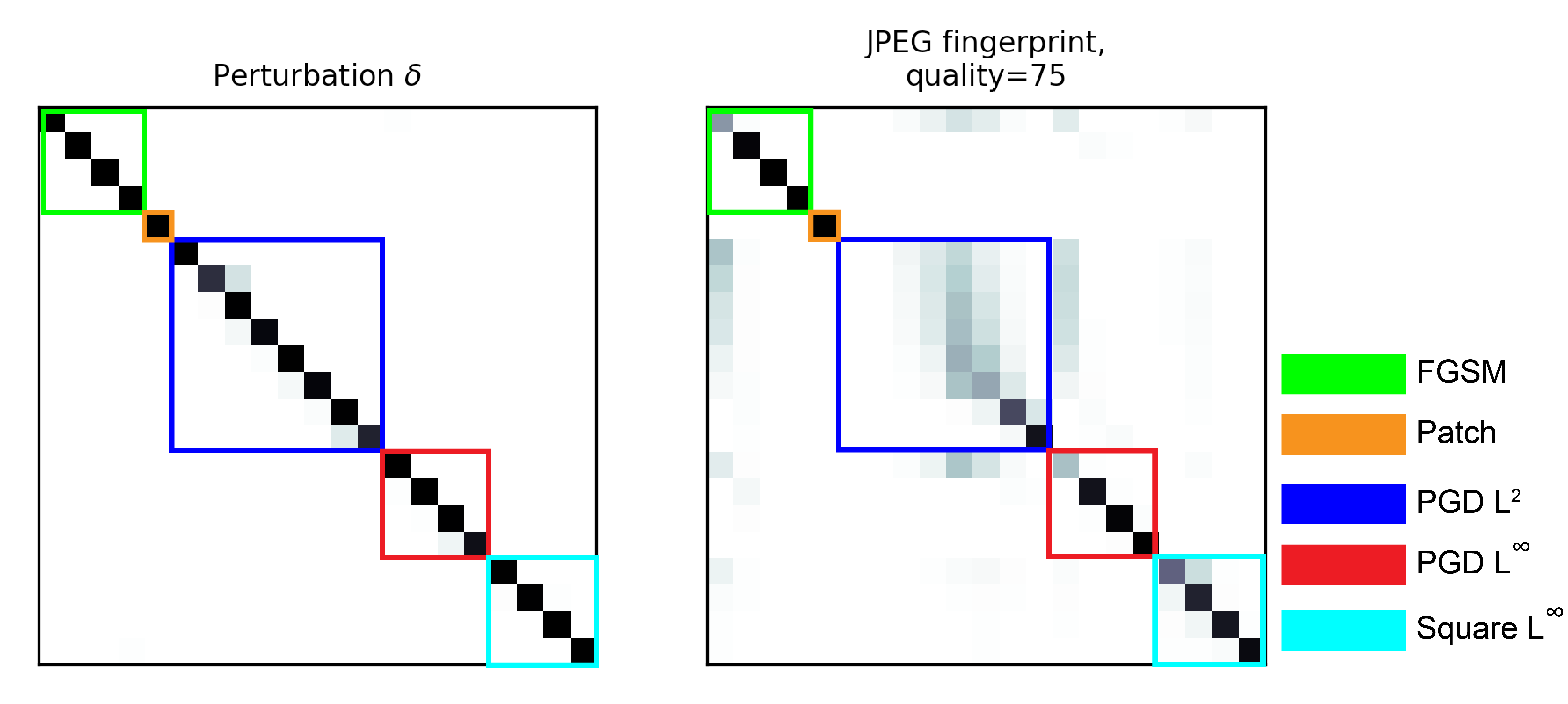}
\caption{Confusion matrices for ResNet50 models trained on dataset with more $\epsilon$ values for PGD-$L^2$ attack.
        As in Figure~\ref{fig:confmat-1}, rows are ground truth labels, columns are predicted labels, 
         and grayscale intensity in each square indicates probability of predicting each label, normalized within rows. Colored rectangles indicate attack algorithms.
         $\epsilon$ values are sorted within attack algorithm to increase from 
         top to bottom and from left to right.
         Each plot is generated from predictions of one training replicate of a ResNet50 model. 
         In general, results were similar across replicates and so a representative example is shown.
}
\label{fig:confmat-more-eps}
\end{figure}

Again we saw that 
when given the true perturbation $\delta$, 
networks could classify this expanded dataset quite well, 
achieving 97.18\% $\pm$ 1.96\% accuracy.
In contrast, we saw a drop in accuracy for models 
trained on JPEG fingerprints (quality=75), 
from 85.05\% $\pm$ 0.83\% we saw before to 72.11\% $\pm$ 1.65\% 
on this expanded dataset with more $\epsilon$ values 
that were ``closer'' to each other.
We generated confusion matrices for these models 
and indeed saw that for the PGD-$L^2$ class, 
the models trained on JPEG fingerprints 
tended to misclassify all but those 
from attacks generated using the largest $\epsilon$ values.
These confusion matrices are shown in Figure~\ref{fig:confmat-more-eps}.

Taken together, our results provide 
positive evidence 
that it is possible to classify adversarial examples 
according to attack algorithm, 
but that a purely supervised approach 
may be challenged 
when faced with the task of estimating continuous parameters 
like the value of $\epsilon$ bound used with an attack.

\subsection{Additional analysis}

Finally, we carried out additional analyses
to test possible alternate explanations for our results. 

\subsubsection{Image quality}

First we asked whether different attacks 
might have different effects on image quality. 
If so, this could serve as a form of data leakage, 
where the network simply learns to classify an image 
by the amount of noise in it.
To test for this possibility, 
we generated 2-D plots of the mean square error (MSE) 
versus the structural similarity index metric (SSIM) 
for each attacked image, compared to the benign image before attack,  
as shown in Figure\ref{fig:mse-v-ssim}. 
SSIM declined expontentially as MSE increased, 
which is perhaps not surprising, 
but we note that these metrics are not 
necessarily tightly linked; 
SSIM was specifically designed 
as a sensitive perceptual measure 
that detects changes in image quality 
that MSE does not take into account, 
while MSE can increase greatly 
even due to changes that the human eye would not detect 
(e.g. shifting the entire image one pixel in one direction) 
\cite{wangMeanSquaredError2009}.
The important thing to notice here is that 
we did not see any obvious clustering of attack 
according to these two values. 
While this does not let us rule out the possibility 
of data pollution by image quality, 
we felt such an explanation was less likely given this result, 
and so we moved on to consider another possible explanation.

\subsubsection{Distribution of class labels produced by untargeted attacks}

\begin{figure*}
     \begin{subfigure}[b]{0.3425\textwidth}
         \centering
         \includegraphics[scale=0.5]{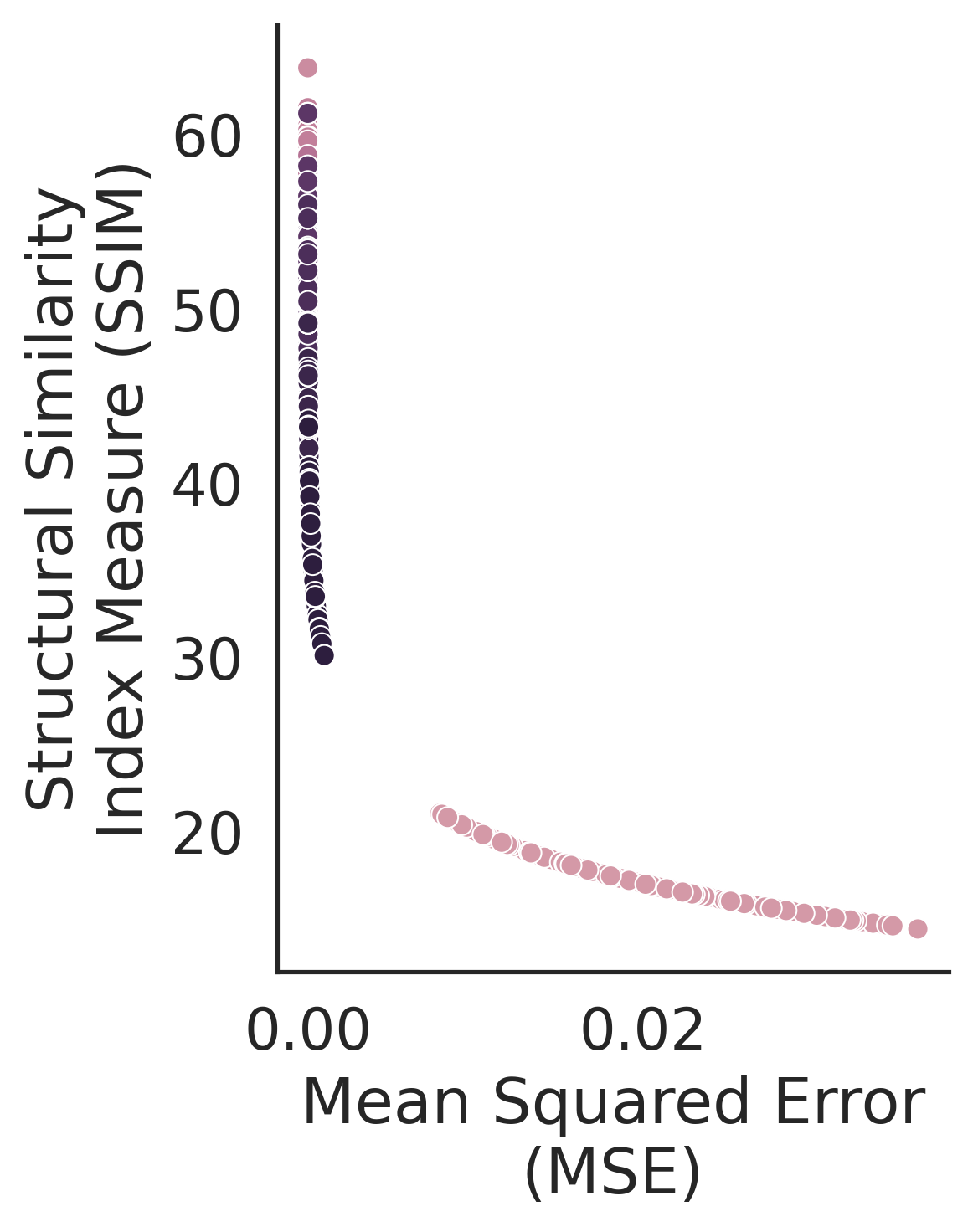}
         \caption{Mean Squared Error (MSE) vs Structural Similarity Index Metric (SSIM).
Hue indicates attack class label. 
Measurements are taken from attacked images, in comparison with original benign images.}
         \label{fig:mse-v-ssim}
     \end{subfigure}
     \hfill
     \begin{subfigure}[b]{0.6425\textwidth}
         \centering
         \includegraphics[scale=0.6]{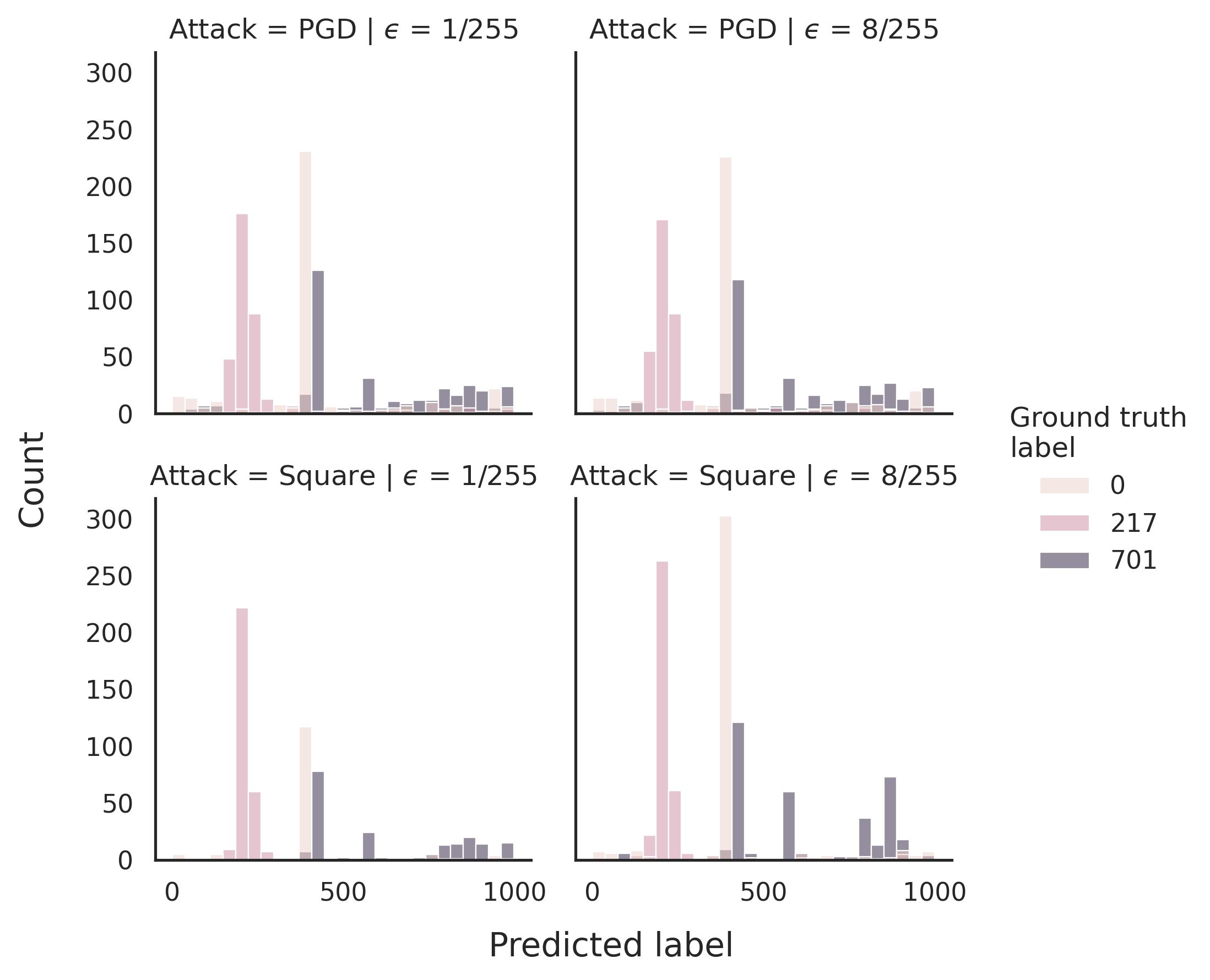}
         \caption{Distribution of labels produced by untargeted attacks. 
         Distributions of predicted labels are shown for three ground truth classes, 
         for two different attacks (rows) and two different $\epsilon$ values (columns)}
         \label{fig:attack-label-distrib}
     \end{subfigure}
        \caption{Analysis of classification results}
        \label{fig:analysis-classification}
\end{figure*}

A second alternative explanation we considered 
was that different attacks might consistently generate 
specific labels for untargeted attacks, 
and this could provide a shortcut that the network would learn. 
I.e., an untargeted PGD-$L^\infty$ attack with $\epsilon=4$ might tend to 
convert the ``fish'' class into ``airplane'', 
whereas the untargeted Square-$L^\infty$ attack 
might tend to convert the same ``fish'' class into ``truck''. 
To test for this possibility, 
we plotted the distribution of targeted labels produced by each attack 
for PGD-$L^\infty$ and Square-$L^\infty$, for all classes and all epsilon sizes.
In Figure~\ref{fig:attack-label-distrib} we show these results. 
For readability, the distribution fro only three classes is shown. 
This analysis did not produce evidence 
suggesting that different attacks produce 
different distributions of labels that a model might be able to learn. 
In fact, we observed the opposite, 
the distributions appear to be quite similar 
across attack types and across $\epsilon$ values. 
For example, in the upper left panel, 
the PGD-$L^\infty$ attack was most likely to convert images 
with ground truth label 0 (``tench'') 
to label 389. 
This was the case in all other panels as well, 
and for all other classes, 
although it was also clear that the label distributions produced 
by attacks on some classes were higher entropy than others 
(compare for example the distributions produced for class ``0'' 
with the distributions for class ``701'' shown in Figure~\ref{fig:attack-label-distrib}). 
This result suggests that the distributions of labels generated by attacks 
are largely a function of the decision boundaries learned by the model under attack, consistent with previous work \cite{goodfellowExplainingHarnessingAdversarial2015,warde201611}, 
and strengthens our claim 
that deep neural networks trained on the perturbations or fingerprints 
are learning to classify the attacks and epsilon sizes, 
not the targeted label, since the latter can vary highly within a ground truth class.

\section{Discussion}

We investigated whether it is possible to reverse engineer attacks, 
in part because of the difficulties faced in developing defenses against them.  
More specifically, we formulated reverse engineering as a problem of classification 
with supervised learning methods. 
We found that we were able to classify adversarial examples according to the 
attack algorithm and the size of the epsilon bound used, 
given the true perturbation added to the attack. 
Classifying the attacked images themselves was not sufficient, 
although we observed that it was possible for attacks with large perturbations 
(e.g., large values of the bound $\epsilon$ for bounded attacks). 
Additionally, we tested whether we could ``fingerprint'' attacked images 
to obtain an estimate of the perturbations added by attack algorithms. 
We showed that the well established and widely available JPEG algorithm 
can be used to provide a good estimate of the perturbation added 
by a wide range of attacks, and that neural network models trained on these fingerprints 
achieved $\sim$85\% accuracy.

These results are not without caveats. 
Our study focuses on reducing the problem to its simplest form, 
and so we have not for example tested our approach on adaptive adversarial attacks, 
and we have not tested whether we can modify adaptive attacks 
to render the task of classifying adversarial examples difficult. 
In spite of these limitations, 
the results presented here are still important, 
directly demonstrating for the first time 
that classifying perturbations according to attack is possible, 
and providing a strong baseline for future work 
for more sophisticated methods of reverse engineering attacks.

\subsection{Future work}

We identify several directions that future studies can take based on our results.
The first would be to combine classification with regression 
to better estimate parameters such as the value of $\epsilon$ used as a bound. 
As we observed, deep neural network models are quite capable of classifying 
attack algorithms given the true perturbation,
but it is more difficult to classify the values of parameters such as $\epsilon$. 
A method to achieve both might be to combine classification 
with a regression loss, 
in the same way that object detection models 
apply classification loss to the labels predicted for bounding boxes 
and regression loss to the coordinates of the bounding box. 
This would allow for prediction of the continuous variable of the epsilon bound 
and could be extended to other parameters such as the number of steps.

Our results on fingerprints also make a strong case 
for a learnable, attack-agnostic method for estimating the perturbation $\delta$. 
This data-driven approach may seem obvious 
to researchers with a deep learning mindset,  
but the clear differences we observed 
between signal processing algorithms 
suggest that careful study 
may incorporate appropriate biases or constraints 
into fingerprinting models 
that improve performance. 
In particular the superior performance of JPEG 
compared to compressed sensing 
suggests that 
a neural network model for estimating fingerprints 
may benefit from integrating 
the perceptual components 
of the JPEG algorithm 
into its architecture. 
Previous work has shown 
that the JPEG algorithm can be implemented as a neural network \cite{shinJPEGresistantAdversarialImagesa}, 
and recent studies of adversarial attacks 
have suggested that 
algorithms tend to place energy 
in specific channels of the color space 
used by the JPEG algorithm \cite{pestanaAdversarialPerturbationsPrevail2020,pestanaAdversarialAttacksDefense2021}.

A last consideration for future studies will be 
how to deal with unknown attacks 
that are not contained within the dataset. 
This would need to be studied with open-world 
and open-set classification problem formulations. 
A natural starting point would be modified loss functions 
such as the entropic loss proposed by \cite{dhamijaReducingNetworkAgnostophobia2018}. 
Note that the derivation of the entropic loss assumes 
a neural network with a fully connected layer 
without a bias term just before the final output layer, 
which may limit its applicability with state-of-the-art neural network models 
for single-label image classification such as ResNet,
although practically speaking one can just add such a layer to a ResNet model 
as \cite{dhamijaReducingNetworkAgnostophobia2018} do in their experiments.

\section{Conclusion}

We have shown that the exquisite sensitivity of deep neural networks 
to adversarial examples can be converted from a bug into a feature. 
Our results are consistent with the idea that 
deep neural network models can classify adversarial examples 
according to attack algorithm. 
These findings set the stage for future work on reverse engineering adversarial attacks.

\bibliographystyle{IEEEtran}
\bibliography{IEEEabrv,reverse-engineer}

\end{document}